\documentclass[final]{article}

     \usepackage[nonatbib]{tackling_climate_workshop_style}

\usepackage[utf8]{inputenc} 
\usepackage[T1]{fontenc}    
\usepackage{hyperref}       
\usepackage{url}            
\usepackage{booktabs}       
\usepackage{amsfonts}       
\usepackage{nicefrac}       
\usepackage{microtype}      
\usepackage{todonotes}
\usepackage{cleveref}

\title{Improving accuracy and convergence of federated learning edge computing methods for generalized DER forecasting applications in power grids}

\author{%
  Vineet Jagadeesan Nair \\
  Department of Mechanical Engineering \\
  Massachusetts Institute of Technology \\
  Cambridge, MA 02141 \\
  \texttt{jvineet9@mit.edu} \\
  \And 
  Lucas Pereira \\
  ITI, LARSyS \\
  Instituto Superior Técnico, University of Lisbon \\
  Av. Rovisco Pais 1, 1049-001 Lisboa \\
  \texttt{lucas.pereira@tecnico.ulisboa.pt}}

\begin{document}

\maketitle

\begin{abstract}
This proposal aims to develop more accurate federated learning (FL) methods with faster convergence properties and lower communication requirements, specifically for forecasting distributed energy resources (DER) such as renewables, energy storage, and loads in modern, low-carbon power grids. This will be achieved by (i) leveraging recently developed extensions of FL such as hierarchical and iterative clustering to improve performance with non-IID data, (ii) experimenting with different types of FL global models well-suited to time-series data, and (iii) incorporating domain-specific knowledge from power systems to build more general FL frameworks and architectures that can be applied to diverse types of DERs beyond just load forecasting, and with heterogeneous clients.
\end{abstract}

\section{Introduction and motivation}

This project aims to improve the accuracy of edge computing methods like Federated Learning (FL) for applications in power systems operation and optimization. Specifically, we tackle the challenge of accurate forecasting to appropriately coordinate Distributed Energy Resources (DER) in medium to low-voltage distribution grids. Our initial focus will be on accurately predicting both short and long-term forecasts of electricity consumption from loads as well as generation from renewables. In addition to forecasting these aggregated quantities at the secondary and primary feeder levels, we will also study individual forecasts at the level of each home or building. Such DER forecasts are crucial for grid operators, utilities, and other entities to operate and optimize power grids. Better forecasts can also help achieve more efficient scheduling and dispatch in order to lower operating costs and electricity tariffs, reduce network power losses, and increase grid reliability and resilience. The need for accurate forecasting is becoming even more acute with the increasing penetration of DERs which introduce more uncertainty, intermittency, and variability. FL is also well-suited to this task since it preserves the data privacy and security of individual consumers and DER owners (prosumers) unlike conventional centralized machine learning approaches.

DER forecasts span a multitude of different kinds of devices. This includes consumption from both fixed or inflexible and flexible loads like electric vehicle chargers, battery storage, and Heating, Ventilation, and Air Conditioning (HVAC) systems. In addition, grid operators also need to predict outputs from renewables such as rooftop solar Photovoltaic (PV) panels. Prior works have looked at applying FL for load and DER forecasting \cite{VenkataramananDERLearning}, but there are still several limitations and challenges remaining that need to be addressed. These include (i) poorer performance of FL in terms of prediction accuracy when compared to traditional centralized learning methods, especially when applied to data that is not independently and identically distributed (IID) across different FL clients, and (ii) long training times for the aggregated global model to converge due to diverse weights and parameters across clients \cite{Yang2019FederatedLearning}. In our forecasting applications, these clients would correspond to different groups of independently-owned DERs distributed across the grid. 

\section{Machine learning methods}

This work aims to address these challenges by applying variants and extensions of FL such as clustered FL and model personalization or fine-tuning in order to improve forecasting accuracy and achieve faster convergence while still preserving privacy and reducing communication requirements relative to centralized machine learning (ML) techniques. In particular, methods like hierarchical clustering (HC) \cite{Briggs2020FederatedData} and Iterative Federated Clustering Algorithm (IFCA) \cite{GhoshAnLearning} recently proposed in the literature have been shown to improve validation and test set prediction accuracy while also achieving faster convergence times by reducing the communication costs and iterations needed to produce a good global inference model. Some of the above client clustering techniques have been applied to niche load forecasting applications such as short-term residential demand forecasting for individual houses and aggregate loads \cite{Tun2021FederatedAggregation, BriggsFederatedForecasting, Gholizadeh2022FederatedForecasting}. However, to our knowledge, such methods have not yet been extended to more general DER forecasting applications as well as longer-term predictions, both of which may introduce much more data heterogeneity across different clients and nodes as well as potential spatio-temporal distribution shifts over time.

In addition to the above extensions, other variants of FL have been proposed that aim to improve the model accuracy through local model personalization and fine-tuning \cite{BriggsFederatedForecasting}, meta-learning \cite{Alp2021DebiasingTraining} and dynamic regularization to better align the optimality of local device-level solutions and the global model \cite{Alp2021FederatedRegularization}. Other studies have developed frameworks aimed at training heterogeneous local models that can still be stably aggregated to produce sufficiently accurate global models \cite{Diao2020HeteroFL:Clients}. However, such methods have not yet been tested or validated for power grid applications like load or DER forecasting. 

In addition to experimenting with improved versions of FL, we also plan to incorporate the use of deep neural network models well-suited to time series data such as long short-term memory networks (LSTM), recurrent and convolutions neural nets for the global and local models, and rigorously evaluate their performance in the FL setting in terms of accuracy as well as communication requirements and sample efficiency. In terms of training the FL client models, we also plan to utilize momentum-based and other accelerated stochastic gradient descent methods for faster convergence \cite{Ruder2016AnAlgorithms}. 

Our goal is to combine the above tools to build a general framework that can be used to accurately forecast different types of DERs (not only loads) with (i) lower prediction errors for individual forecasts relative to local learning and (ii) lower errors for aggregated forecasts relative to centralized ML, while still retaining data privacy, sample efficiency, and computational speed. In addition to predicting the baseline power injections (generation or load) from these DERs, we also aim to forecast their associated upward or downward flexibilities that they can offer to grid operators. Accurately capturing this DER flexibility information is becoming increasingly critical to efficiently and reliably run transmission and distribution grids under the new paradigm of distributed generation/storage, flexible demand response capabilities, and more frequent extreme weather events. In order to build such generalizable and high-fidelity models, we will need to combine the latest advancements in ML with domain-specific knowledge such as load models \cite{Arif2018LoadReview}, socioeconomic and behavioral factors for DER owners, weather-based renewables output forecasting (e.g. solar irradiance for PV output), and physics-based grid constraints for storage, generator ramp rates, etc. Another example is the distinct types of load flexibilities between shiftable versus curtailable versus non-interruptible loads. Incorporating such factors would help us build physics-informed models that are more well-suited to power system applications.

Finally, a more open-ended aspect of the project will examine and survey approaches to further enhance the security and privacy of FL methods, especially in the context of cyber-physical threats in power grids and the risk of exposing the data characteristics of individual clients and DERs. Here, we hope to explore the use of methods like differential privacy using noisy signals and secure multi-party computation, particularly for DER forecasting at an aggregate level. In doing so, we can also evaluate and consider the tradeoffs between privacy protection and prediction accuracy, and/or the duality between model performance (in terms of accuracy and computational speed) versus robustness. 

\section{Climate change and real-world impacts}

Improved time series forecasting for loads and other distributed energy resources would help us coordinate these resources more effectively in real-time to operate power grids more efficiently and optimally. More accurate forecasts could also help grid operators plan better to maintain system stability, and improve system reliability and resilience, while also reducing network losses, lowering operational costs, and achieving other key objectives like voltage regulation and frequency control. The improved observability and controllability of distribution grids resulting from better DER forecasts would also facilitate easier coordination with the main transmission grids and wholesale markets - since grid operators currently don't have much visibility over the distribution system downstream. 

Such methods would help accelerate the deployment and more seamless integration of DERs into distribution grids in order to increase the penetration of renewables as well as other supporting technologies like demand response, batteries, and electric vehicles. All of this will help with the rapid decarbonization of the electric power sector as well as other related sectors like transportation, heat, and heavy industry that will eventually be electrified as well in the near future. In addition to helping drive down emissions, this would also help improve affordability by lowering energy costs for customers, and promote climate justice by increasing access to clean energy. 

\begin{figure}[!h]
    \centering
    \includegraphics[width=\textwidth]{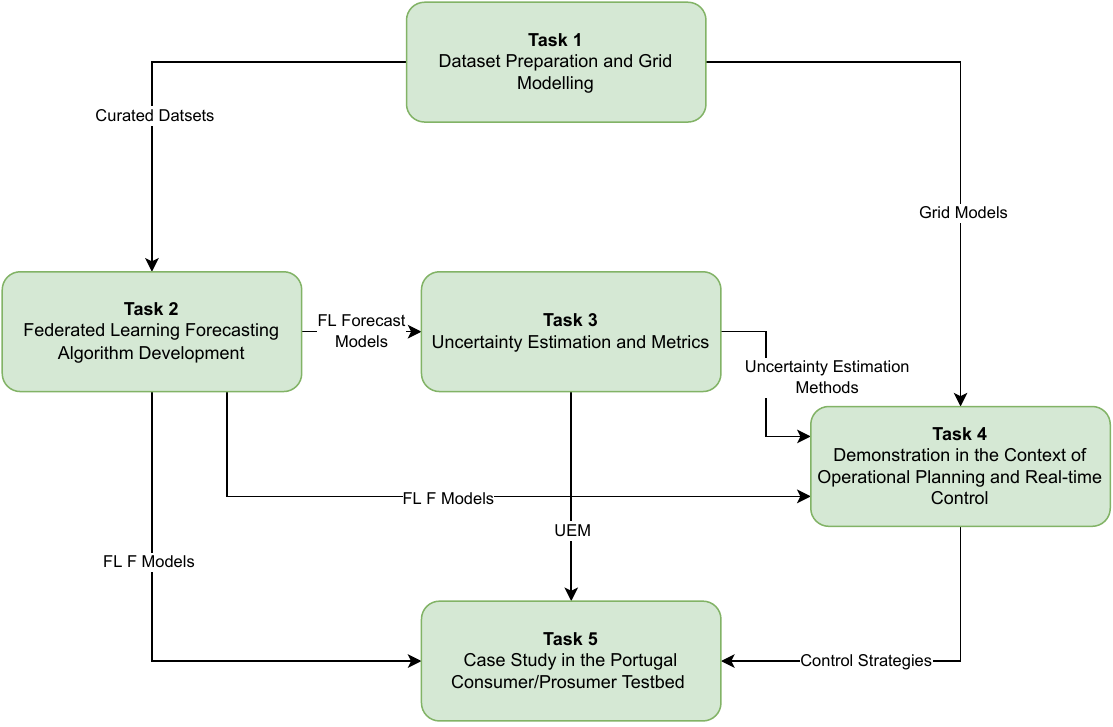}
    \caption{Summary of key tasks and activites proposed for this research project.}
    \label{fig:activities}
\end{figure}

We aim to demonstrate these impacts by exhaustively validating our proposed FL frameworks on large-scale, realistic data sets to demonstrate the effectiveness of FL over traditional time-series forecasting methods. We aim to start with standard IEEE power grid test cases \cite{Arritt2010TheFeeder} and then extend to more realistic scenarios using real smart-meter utility data \cite{Bu2019AData} and other sources for DER data like Pecan Street \footnote{\url{https://www.pecanstreet.org/}}. In addition, we also hope to extend our testing beyond the US to look at other regions with high DER penetration like the EU. The flowchart in \cref{fig:activities} shows the mains tasks that will be undertaken to complete this proposal.

\bibliographystyle{IEEEtran}
\bibliography{references}

\end{document}